\documentclass[letterpaper, 10 pt, conference]{ieeeconf}

\IEEEoverridecommandlockouts

\usepackage{times}
\usepackage{epsfig}
\usepackage{graphicx}
\usepackage{amsmath}
\usepackage{gensymb}
\usepackage{amssymb}
\usepackage[]{multirow}
\usepackage{float}
\usepackage{graphicx}
\usepackage{subcaption}

\usepackage[pagebackref=true,breaklinks=true,letterpaper=true,colorlinks,bookmarks=false]{hyperref}

\begin{document}
\title{Any Motion Detector: Learning Class-agnostic Scene Dynamics from a Sequence of LiDAR Point Clouds}

\author{Artem Filatov \\
Yandex \\
{\tt\small filatovartm@yandex-team.ru}
\and Andrey Rykov \\
Yandex\\
{\tt\small a-rykov@yandex-team.ru}
\and Viacheslav Murashkin \\
Google \\
{\tt\small vmurashkin@google.com}
}

%%%%%%%%% TITLE
\maketitle
%\thispagestyle{empty}

%%%%%%%%% ABSTRACT
\begin{abstract}
Object detection and motion parameters estimation are crucial tasks for self-driving vehicle safe navigation in a complex urban environment. In this work we propose a novel real-time approach of temporal context aggregation for motion detection and motion parameters estimation based on 3D point cloud sequence. We introduce an ego-motion compensation layer to achieve real-time inference with performance comparable to a naive odometric transform of the original point cloud sequence. Not only is the proposed architecture capable of estimating the motion of common road participants like vehicles or pedestrians but also generalizes to other object categories which are not present in training data. We also conduct an in-deep analysis of different temporal context aggregation strategies such as recurrent cells and 3D convolutions. Finally, we provide comparison results of our state-of-the-art model with existing solutions on KITTI Scene Flow dataset.
\end{abstract}

\section{Introduction}
Modern object detectors do great at detecting objects of classes which are present in training dataset \cite{zhou17voxelnet}. However, besides common categories of traffic participants like cars,  pedestrians, and cyclists many other moving objects of different shapes and sizes may occur on the road, e.g., dogs, other animals or moving oversized construction machines. Such a complex environment poses strong requirements to the generalizability of the perception system to the previously unseen object classes.

\begin{figure}[ht]
    \centering
    \includegraphics[width=0.8\linewidth]{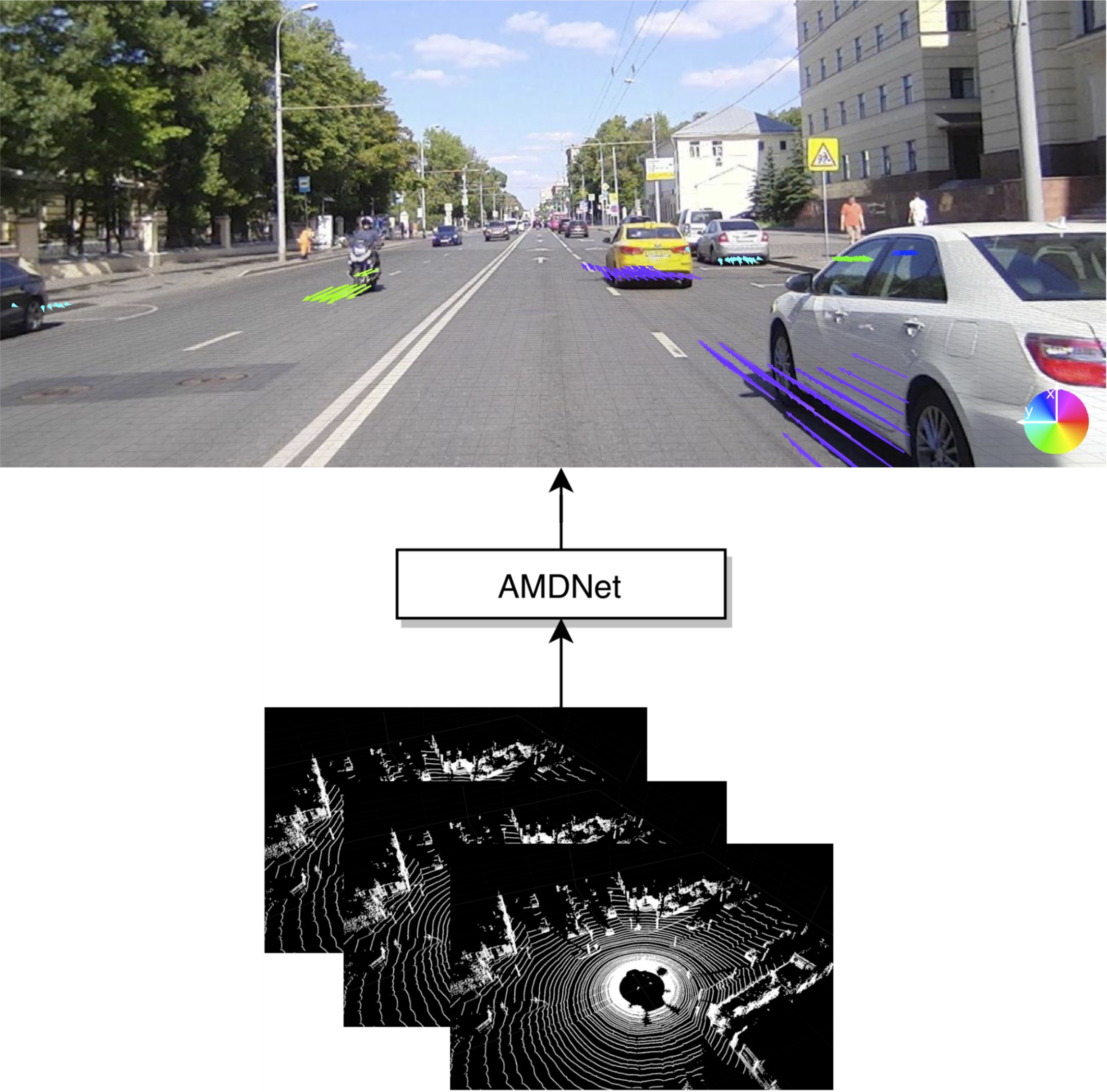}
    \caption{AMDNet takes point cloud stream as input and estimates bird's-eye view 2D scene motion. Camera image is used for visualization purpose only.}
    \label{fig:pipeline}
\end{figure}

Although the understanding of what kind of traffic participants are present on a road is fundamental, accurate scene dynamics estimation is a key for a self-driving vehicle safe navigation in a complex urban environment. One of the widely used approaches to scene dynamics prediction is object trajectory extrapolation using estimated velocity and acceleration. To estimate object motion parameters one needs to incorporate temporal context into the object detection model and analyze several consecutive frames. There also exists a number of neural network based approaches for object detection and velocity estimation task which are trained in an end-to-end fashion \cite{luo2018faf}.

The self-driving vehicle uses the number of different sensors to understand its surrounding environment. One of the most commonly used sensor in self-driving perception system is LiDAR. Scanning environment with laser beams, LiDAR produces 3D point cloud representing distances to the nearby obstacles around the self-driving vehicle. Such point cloud is successfully used in 3D object detection and recognition algorithms \cite{zhou17voxelnet,qi2017frustum}.

A self-driving vehicle captures data from sensors mounted on itself which poses an issue of ego-motion compensation in sensor data for temporal context aggregation process and estimation of surrounding objects' motion parameters.

In this work, we propose the Any Motion Detector (AMDNet) - a novel real-time end-to-end architecture for class-agnostic scene dynamics estimation from a sequence of LiDAR point clouds. The main contributions of our work are as follows:
\begin{itemize}
    \item we propose a new real-time architecture for class-agnostic motion detection in a complex urban environment based on consecutive LiDAR point clouds;
    \item we introduce differentiable ego-motion compensation layer for temporal context aggregation;
    \item we conduct a quantitative analysis of alternative approaches for temporal context aggregation;
    \item we compare our state-of-the-art model with existing solutions on the open KITTI Scene Flow dataset \cite{sceneflow}.
\end{itemize}

\section{Related Work}

Due to the success of deep learning in object detection for camera images, similar approaches were applied to 3D object detection in LiDAR point clouds represented as bird's-eye view tensors. Most of the approaches used voxelization and some hand-crafted features for each voxel \cite{li16velofcn, chen16mv3d, yang2018pixor}. More recent works utilize ideas from \cite{qi2017pointnet} and \cite{qi2017pointnet++} to obtain learnable voxel representations. We believe that such an approach is beneficial for the task of velocity grid estimation, for this reason we base point feature extraction layers of our model on the method proposed in \cite{zhou17voxelnet}.

A natural way to improve object detection or to solve motion estimation tasks is to add temporal context to the input. \cite{luo2018faf} stacks a sequence of point clouds with a fixed length in a 4D tensor and use 3D convolutions to accumulate time information. Though this approach is fast enough for real-time inference, as the length of the sequence is limited, the network cannot capture long-time dependencies such as part-time occluded objects. Moreover, to estimate the speed of an object \cite{luo2018faf} uses some prediction-correction framework based on the future trajectory estimations made by their network from past frames and current detections. Such an approach seems less robust compared to our end-to-end trained system as a missed detection, or wrong future estimation automatically leads to the wrong speed estimate.

Another way to obtain objects' speed is to estimate the scene flow. \cite{wang2018contconv} proposes a novel continuous convolution operation and uses features from 2 sequential point clouds to predict pointwise speed. \cite{liu2018sceneflow} successfully applies \cite{qi2017pointnet++} to 2 sequential point clouds and also predicts pointwise speed. Though their approach gives promising results, it does not run in real-time.

\cite{behl2018pointflownet} use ideas from \cite{zhou17voxelnet} to obtain point cloud features and combine information from 2 sequential clouds to predict ego-motion, cars' detection, speed in bird's-eye view grid and rigid body motion for cars.

\section{Problem Formulation}

In this work, we develop an architecture for dynamic object detection and motion parameters estimation by the sequence of 3D point clouds. The input to our model is a sequence of point clouds $\{P_i\}_{i=1}^p$ and corresponding transforms $T_i$ from the sensor's local coordinate system at time $t_{i}$ to world coordinates. The output of the model is the bird's-eye view grid for dynamic/static segmentation of each cell and 2D velocity of each cell in the XY plane. The X-axis is oriented along the ego-vehicle movement direction, Y-axis is orthogonal to X, and Z-axis is directed vertically. We also consider only the urban setting where the most dynamics is in the 2D plane, and we argue that velocity along the vertical direction is not as much important and useful for road participants. Figure \ref{fig:pipeline} demonstrates the whole pipeline.

\section{AMDNet architecture}

\begin{figure*}[ht]
\begin{center}
    \includegraphics[width=1.\linewidth]{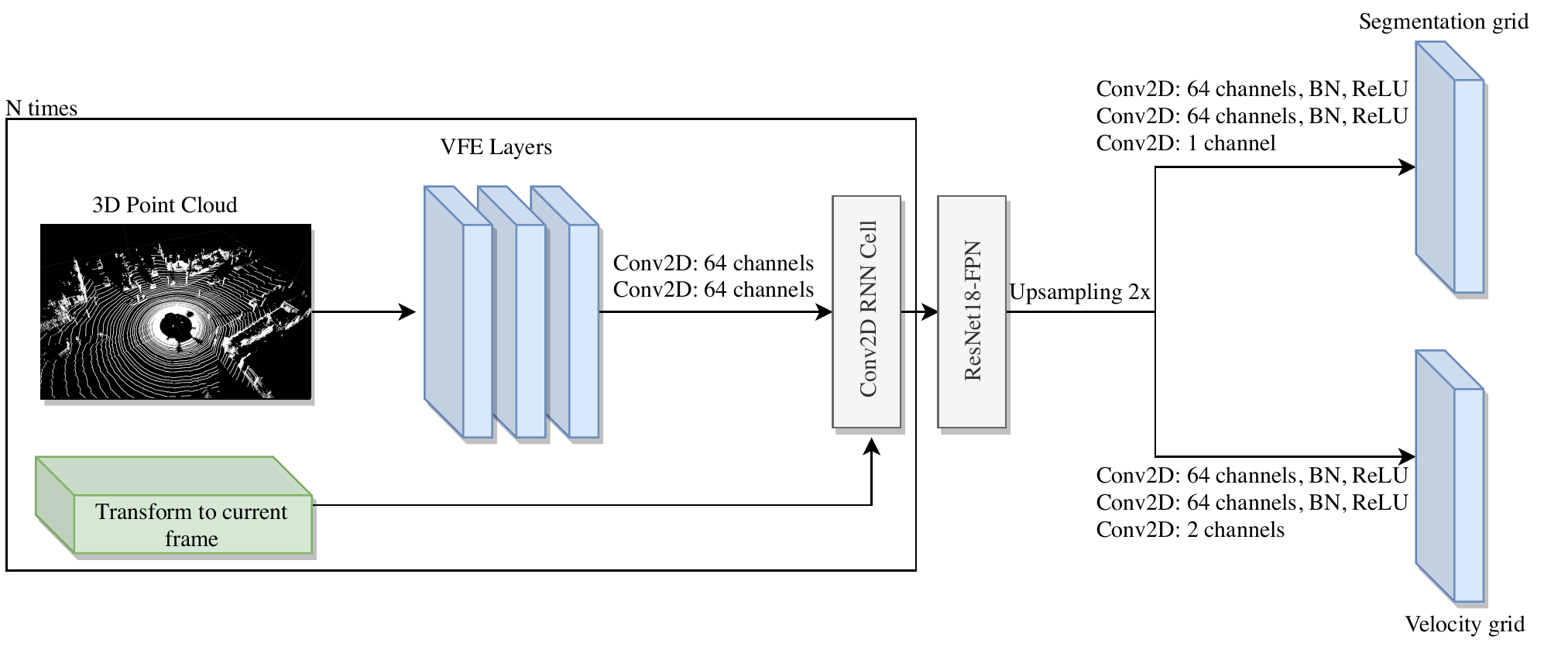}
\end{center}
   \caption{AMDNet architecture. Each raw point cloud in a sequence is processed with a stack of Voxel Feature Encoding layers. Afterward, temporal context aggregation is performed using convolutional RNN cell with ego-motion compensation layer applied to the hidden state. The last hidden state is passed through ResNet18-FPN backbone network. Finally, two branches output 2D velocity and dynamic/static segmentation grids.}
\label{fig:arch}
\end{figure*}

We propose the AMDNet: an end-to-end architecture which takes a sequence of point clouds and transforms $T_i$ and outputs bird's-eye view grid motion segmentation and velocity estimation for each cell. AMDNet consists of the following parts: Voxel Feature Encoding layers for point cloud feature extraction, RNN cell with ego-motion compensation layer for time-context accumulation, ResNet18-FPN and two branches that output network predictions. One of the outputs is the 2D velocity grid, and another one is the binary dynamic/static cell segmentation. We use dynamic/static segmentation to mask velocity grid. It helps to eliminate incorrect velocity predictions which can occur in regions without points  due to a specific loss function described in Section \ref{sec:loss}. Figure \ref{fig:arch} presents the whole architecture.

\subsection{Voxel Feature Encoder}
 
To obtain the bird's-eye view feature representation of the point cloud, we used the approach similar to VoxelNet \cite{zhou17voxelnet}. We divide local 3D space into voxels with predefined height, width, and length. Such discretization naturally introduces mapping $D$ from point cloud local coordinate system to tensor coordinates. After discretization, we group point cloud according to voxels membership. After that, we extend each coordinate with the relative coordinate inside the voxel:
\begin{equation}
    \label{eq:local_point_features}
    \begin{aligned}
    p_i = &(x_i, y_i, z_i) \rightarrow \\
    &\tilde{p_i} = (x_i, y_i, z_i, x_i - x_{\mathrm{cen}}, y_i - y_{\mathrm{cen}}, z_i - z_{\mathrm{cen}})
    \end{aligned}
\end{equation}
 where $\bf{x}_{\mathrm{cen}}$ corresponds to voxel's center coordinate. Then we process points in each voxel independently with Voxel Feature Encoding layer \cite{zhou17voxelnet} multiple times. After that, we make max-pooling over all points in the voxel to obtain a single feature vector.
 
 Unlike the original paper, we do not apply 3D convolutions and instead stack all the feature vectors along the vertical axis to obtain 3D tensor which describes the space around the observer. We also exploit the sparseness of the voxelized representation and proceed with $1\times 1$ convolutions over non-zero columns to squash channel dimension that reduces inference time. Finally, we apply two convolutional layers to take into account dependencies between neighboring columns.

\subsection{Ego-Motion Compensation Layer}

To estimate the motion of dynamic objects in the scene we need to aggregate temporal context. Voxel Feature Encoder outputs tensor representations of consequent point clouds, but each of them is in its local coordinate system. If one naively aggregates such representations using RNN or 3D convolutions, the model would be forced to learn ego-motion of the observer implicitly. We believe that such setting lowers the quality of the resulting motion estimation. However, as in a self-driving setting localization of the ego-vehicle is a crucial part of the pipeline, ego-motion is often known precisely. Given transforms from the local coordinate system to the world coordinates from localization module it is quite natural to compute transforms $T_{i-1, i}$ between two consecutive local coordinate systems. Namely, $T_{i-1, i} = T^{-1}_{i} T_{i-1}$ where $T_{i}$, $T_{i-1}$ are transforms from local to world coordinates.

An obvious way to compensate ego-motion in temporal context aggregation process is to transform all previous clouds in the sequence to the local coordinates of the last one. While this approach does not have discretization error, we argue that it has limited application in the real-time system because of the slow inference. Raw point clouds transformation requires recomputation of feature tensors for the whole sequence which is expensive regarding the time. We propose to use a simple but quite effective technique regarding quality and inference time that is able to reuse already computed feature tensor from the previous timestamp. Given some feature tensor for $i-1$ timestamp $H_{i-1}$, transform $T_{i-1, i}$ and mapping $D$ from local coordinate system to tensor coordinate systems our ego-motion compensation layer computes the following tensor $\hat{H}_{i-1} = W(H_{i-1}, T_{i-1, i}, D)$ where $W$ is a well-known warp operation that performs tensor shift and rotation using bilinear interpolation. This operation is fast and does not introduce significant overhead. Moreover, it is important that such operation is differentiable with respect to $H_{i-1}$, therefore, it can be used as a layer in a neural network allowing to train it using standard backpropagation. After application of ego-motion compensation layer $\hat{H}_{i-1}$ is spatially aligned with the consecutive point cloud tensor representation.

\subsection{Temporal Context Aggregation}

\begin{figure}[ht]
    \centering
    \includegraphics[width=1.\linewidth]{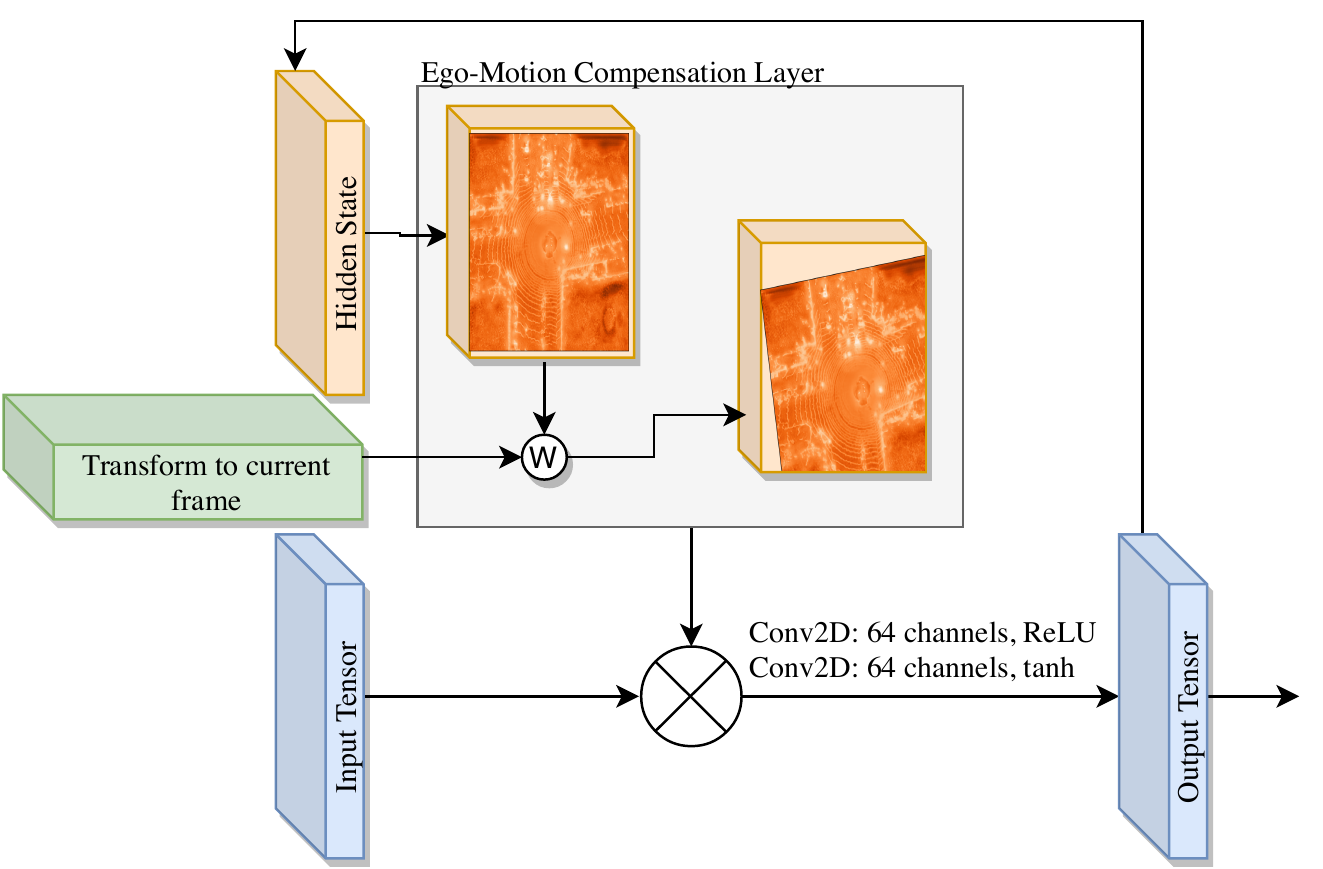}
    \caption{RNN cell with ego-motion compensation layer. We take a hidden state and apply an affine transform to it using the odometry between the previous and current frames.}
    \label{fig:rnn_cell}
\end{figure}

To aggregate temporal context, we use convolutional RNN cell with ego-motion compensation layer for previous hidden state. The equations of the RNN cell are shown below: 

\begin{equation}
    \label{eq:rnn_cell}
    \begin{aligned}
        & \hat{H}_{i-1} = W(H_{i-1}, T_{i-1, i}, D) \\
        & \hat{X_i} = \mathrm{Conv2D}([X_i, \hat{H}_{i-1}]) \\
        & H_i = \mathrm{Conv2D}(\hat{X_i})
    \end{aligned}
\end{equation}
where $H_{i-1}$ and $H_i$ are hidden states of consecutive timestamps, $W$ is ego-motion compensation layer, $X_i$ is feature tensor obtained from voxel feature encoder, and $[\cdot, \cdot]$ denotes concatenation along channels dimension. Schematically this cell is presented in Figure \ref{fig:rnn_cell}.

\subsection{Feature Extractor}

After temporal context aggregation, we obtain the feature tensor which stores the information about the past. We apply ResNet18 Feature Pyramid Network \cite{resnet, lin16fpn} to learn the high-level features.
Then we pass the features through two distinct branches to get dynamic/static segmentation and velocity grids. 

\subsection{Loss Function} \label{sec:loss}
The loss function is a weighted sum of two components: velocity loss and segmentation loss:
\begin{equation}
    \label{eq:loss}
    L = L_{\mathrm{vel}} + \alpha  L_{\mathrm{seg}}
\end{equation}

We use smooth $L_1$ loss \cite{fast_rcnn} for velocity grid. Velocity loss is calculated only for those cells that contain information about the object's speed:
\begin{equation}
    \label{eq:vel_loss}
    L_{\mathrm{vel}} = \frac{1}{N_{v_c}}\sum_{c: v_c \text{is known}} \mathrm{SmoothL1}(v_c, \hat{v}_c)
\end{equation}
where $v_c$ is the ground truth 2D speed for the cell with index $c$, $\hat{v}_c$ is the corresponding AMDNet speed estimation and $N_{v_c}$ is the number of cells with known speed for current scene. Computation of the loss only for cells with known speed allows us to avoid propagation of false negative signal for the dynamic objects that are not presented in the ground truth. In Section \ref{seq:experiments} we demonstrate, that AMDNet can generalize well and detects arbitrary scene dynamics.

Our loss for dynamic segmentation grid
is weighted binary cross-entropy:
\begin{equation}
    \label{eq:seg_loss}
    L_{\mathrm{seg}} = -\frac{1}{N}\sum_c (\beta p_c \log(\hat{p}_c) + (1-p_c)\log(1 - \hat{p}_c))
\end{equation}

Low amount of dynamic objects in the scene leads to a high imbalance between dynamic and static cells in ground truth grid. Moreover, as the recall of ground truth dynamic cells segmentation is not perfect, we want AMDNet to be more sensitive to the dynamic part of the segmentation loss. For this reason we set $\beta=100$.

Finally, in Equation \ref{eq:loss} we set $\alpha=5$ as such value of the weight gives the best results on our validation set.

\section{Experiments} \label{seq:experiments}

\subsection{Training Data}

\label{sec:data}

\begin{figure*}[ht]
\begin{center}
    \includegraphics[width=1.\linewidth]{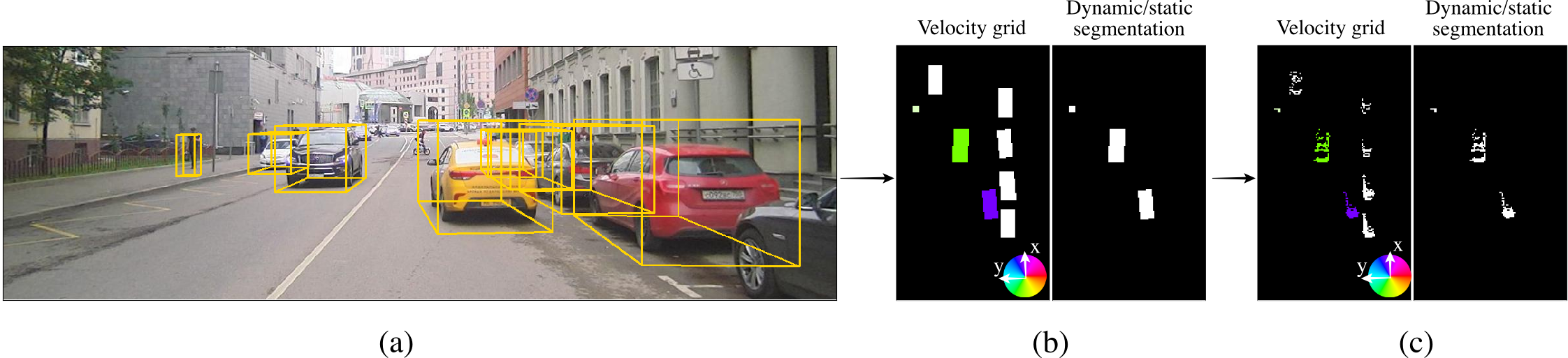}
\end{center}
   \caption{Data preparation pipeline. We project 3D bounding boxes (a) and velocities onto the grid (b). Hue component indicates the velocity direction according to HSV color space. Saturation indicates the absolute velocity value. Refinement step (c) leaves only those cells that contain LiDAR points.}
\label{fig:data}    
\end{figure*}

Currently, there does not exist open dataset which contains labeled velocities for all dynamic objects in the scene in the self-driving setting. We argue that we can use 3D bounding boxes with assigned velocities to obtain segmentation and velocity grid by procedure described in Figure \ref{fig:data}.

Refinement procedure helps us to avoid predictions in the regions where the network does not have enough information. Moreover, such refinement provides more precise grid as the actual geometry of labeled objects is far more complex than simple 3D boxes.

For training, we use our internal dataset with 100k examples. Each example contains a sequence of Velodyne VLP-32C LiDAR point clouds captured at 10Hz. As ego-vehicle has non-zero velocity, we have to perform the same procedure as \cite{kitti} to "untwist" the points applying odometry transformation to corresponding sectors of LiDAR scans. Such transformation allows obtaining point clouds with all the points having the same timestamps.

In order to obtain 3D bounding boxes, we run a pre-trained 3D detector (trained with cars and pedestrians) to collect them for each LiDAR scan. After that, we apply Kalman Filter based tracking algorithm \cite{lipton1998tracker} over bounding boxes to derive trajectories and estimate motion parameters. Finally, we project the object's velocities onto the grid according to the procedure described in Section \ref{sec:data}. The cells of the grid have size $0.2m \times 0.2m$. Then we extract segmentation for dynamic/static objects: if velocity in a given cell is larger than some threshold $\theta$ we mark it as dynamic. Otherwise, we mark it as static. In our experiments we set $\theta=0.8$m/s. During the training we augment the data by random rotation and scaling following the procedure described in \cite{zhou17voxelnet}.

This approach has one significant disadvantage: it includes information only about objects that were labeled by bounding boxes whereas there are a lot of unusual dynamic objects that may occur on the road, e.g. different animals. In Section \ref{seq:experiments} we demonstrate that the model trained on such data is able to generalize to arbitrary dynamics.

\subsection{Training Procedure}
We train all the models with Adam optimizer \cite{adam} for 400k steps. The initial learning rate is $3 \cdot 10^{-4}$, and we make learning rate drop every 70000 steps. We use batch size equal to 4. We also use weight decay with parameter 0.002.

\subsection{Evaluation Datasets and Metrics}
\begin{table*}[ht]
\centering
\begin{tabular}{|l|ccc|ccl|c|}
\hline
\multicolumn{1}{|c|}{\multirow{2}{*}{Method}} & \multicolumn{3}{c|}{with road}                                                             & \multicolumn{3}{c|}{without road}                                         & \multirow{2}{*}{Inference time(ms)} \\ \cline{2-7}
\multicolumn{1}{|c|}{}                        & \multicolumn{1}{l}{3D EPE}  & \multicolumn{1}{l}{3D EPE-dynamic} & \multicolumn{1}{l|}{AP} & \multicolumn{1}{l}{3D EPE}  & \multicolumn{1}{l}{3D EPE-dynamic} & AP     &                                      \\ \hline
ICP-global                                    & \multicolumn{1}{c|}{0.215} & \multicolumn{1}{c|}{}              &                      & \multicolumn{1}{c|}{0.451}  & \multicolumn{1}{c|}{}              &        & \multirow{2}{*}{15}                  \\
ICP-pointwise                                 & \multicolumn{1}{c|}{0.158} & \multicolumn{1}{c|}{0.65}          &       0.781        & \multicolumn{1}{c|}{0.313}  & \multicolumn{1}{c|}{0.63}          &  0.831      &                                      \\
FlowNet3D                                     & \multicolumn{1}{c|}{0.162} & \multicolumn{1}{c|}{0.306}        &       \textbf{0.818}              & \multicolumn{1}{c|}{0.122} & \multicolumn{1}{c|}{0.157}        &  0.938  & 915                                  \\ \hline
AMDNet-RNN                                    & \multicolumn{1}{c|}{0.097} & \multicolumn{1}{c|}{0.13}        &   0.793                  & \multicolumn{1}{c|}{0.099}  & \multicolumn{1}{c|}{0.126}        & 0.936  & 72                                   \\
AMDNet-3Dconv                                 & \multicolumn{1}{c|}{0.097} & \multicolumn{1}{c|}{0.131}        &  0.792                    & \multicolumn{1}{c|}{0.101} & \multicolumn{1}{c|}{0.127}        &  0.931  & 180                                  \\
AMDNet-3Dconv-PCT          & \multicolumn{1}{c|}{\textbf{0.093}} & \multicolumn{1}{c|}{\textbf{0.12}}          &  0.795              & \multicolumn{1}{c|}{\textbf{0.092}}  & \multicolumn{1}{c|}{\textbf{0.114}}     & \textbf{0.94} & 220                                  \\ \hline
AMDNet-RNN-no-odo                                   & \multicolumn{1}{c|}{0.102} & \multicolumn{1}{c|}{0.154}        &   0.83                  & \multicolumn{1}{c|}{0.118}  & \multicolumn{1}{c|}{0.158}        & 0.889  & 70 \\ \hline                
Oracle & \multicolumn{1}{c|}{0.03} & \multicolumn{1}{c|}{0.034} & 0.943 & \multicolumn{1}{c|}{0.028} & \multicolumn{1}{c|}{0.032} & 0.977 & \\ \hline

\end{tabular}
\caption{Comparison results of different methods on KITTI dataset: end-point-error (meters), end-point-error for dynamic points, and average precision for dynamic/static segmentation.}
\label{tab:main}
\end{table*}

We demonstrate the performance of our approach on two datasets: KITTI Scene Flow dataset and synthetic dataset obtained from the simulator. 

\textbf{KITTI Scene Flow dataset} \cite{sceneflow} consists of 200 frames with known disparity and optical flow. Using this information, we extract point clouds for two consecutive frames and compute the ground truth flow between them. The dataset also contains sequences of Velodyne HDL-64E LiDAR point clouds which we use during the AMDNet inference. We evaluate the metrics with the first 130 scenes from the training part.

\textbf{Simulator dataset}. We collected the small dataset with 150 frames from Carla simulator \cite{carla}. We created an environment with houses, pedestrians, bikes, and motorcycles. Dataset consists of LiDAR point clouds simulating Velodyne VLP-32C, transforms between consecutive frames and bounding boxes for all road participants. Whereas LiDAR configuration is close to our training domain, objects look very unnatural. Using this dataset, we demonstrate that AMDNet trained with velocity grids for cars and pedestrians generalizes well to simulated LiDAR and can detect dynamic objects of different shapes.

\textbf{Metrics}. We measure all the metrics pointwise. We measure the quality of the velocity estimation with end-point-error (EPE) \cite{epe}. We also report EPE corresponding only to dynamic points. We mark points as dynamic if the $l_2$-norm of the flow vector is bigger than 0.08m. Following \cite{liu2018sceneflow} we report the average precision (AP) metric for dynamic/static segmentation by predicted velocity value.
We evaluate the metrics in the limits $(0m, 40m)$ - for X-axis and $(-40m, 40m)$ - for Y-axis in vehicle local coordinate system. We also conduct experiments without road surface. In that case, we remove all the points with Z-coordinate lower than $0.1m$. Our method still operates on the full point cloud in that case, and the cropped cloud is used only for metrics computation.

We also measure mean inference time for our models and baselines. We use NVidia GeForce 1080Ti as a GPU accelerator and implement all the nets using TensorFlow 1.12 \cite{tensorflow2015-whitepaper}.

During the inference, we pass five LiDAR point clouds to the AMDNet. Then we take the velocity mask and multiply it by segmentation mask which is binarized by threshold. After that, we project velocity grid onto the point cloud obtained either from disparity for KITTI  dataset or from simulated LiDAR for Carla dataset. Then we multiply the projected velocity by the time interval between two consecutive clouds which is equal to 0.1sec to obtain pointwise flow. We set estimated flow along Z-axis to zero.

\subsection{Baselines}
We compare our method with FlowNet3D \cite{liu2018sceneflow} and ICP \cite{icp}. FlowNet3D is a current state-of-the-art approach solving point cloud 3D flow estimation problem. It takes two 3D point clouds as input and computes flow for each point in the first cloud. We reimplemented the FlowNet3D paper and trained FlowNet3D with FlyingThings3D \cite{flyingthings} dataset as the original paper describes. We use the same hyper-parameters and layers configuration. 

We also conduct experiments with ICP algorithm \cite{icp}. We apply ICP to two consecutive point clouds. Due to a large number of static objects in the scene ICP returns almost identity transform. We also extract pointwise matching from ICP to obtain transforms for each point and use the distance between connected points as flow.

Both baselines operate on point clouds transformed into the coordinate system of the last cloud. For KITTI Scene Flow dataset baselines make inference directly on point clouds obtained through disparity.

\subsection{KITTI Scene Flow Results}

Experiments demonstrate the performance of our method in terms of quality and inference time. Our architecture solves classification and regression problems with very high precision. 
\cite{liu2018sceneflow} mentioned that FlowNet3D performs better without road surface. The results show that our method does not degrade from the presence of the road. This is due to segmentation, which correctly marks the road as a static area.

We conduct experiments with different time-aggregation modules. We demonstrate the results with the 3D convolutions instead of RNN cell. In the AMDNet-3Dconv model, we apply the ego-motion compensation layer to feature tensor for five consecutive point clouds and then apply 3D convolution to stacked tensor. Table \ref{tab:main} shows that 3D convolution does not provide any performance gain but inference time increases dramatically.

To estimate the influence of the ego-motion compensation layer and measure the error which lies inside the discretization error between transforms, we train the AMDNet-3Dconv-PCT with 3D convolution time aggregation module without ego-motion compensation layer. Instead of this, we transform all the point clouds into one coordinate system before passing them to Voxel Feature Encoder. Table \ref{tab:main} shows that AMDNet-3Dconv-PCT models achieve slightly better results due to the absence of the discretization error in transforms. However, computational demands increase dramatically leading to model's disability to operate in real-time.

To emphasize the importance of odometry correction we train AMDNet-RNN-no-odo without ego-motion compensation layer. We also do not transform point clouds before passing them to Voxel Feature Encoder. In order to give high-quality predictions, this model should implicitly take ego-motion into account. Experiments show a drop in the velocity estimation quality. The reason can be the ambiguity between the ego-motion and other objects' motion. At the same time, this model still surpasses the quality of the baselines.

To estimate the error coming from descritized prediction, we propose to measure Oracle metrics. More specifically, we project ground truth flow onto the grid and then project that grid back to points. During the point cloud to grid projection step, if multiple points were projected into one cell, we assign the flow with maximal norm to that cell. Finally, we measure EPE between original flow and reprojected one obtaining the minimal error that we can achieve with the chosen discretization step.

The LiDAR takes full $360 \degree$ scan for 0.1sec. Therefore, the real-time method should work faster than this time. Results demonstrate that our method achieves real-time inference speed. It means that it can be directly used as a part of the autonomous vehicle perception system without any modification. 

%\begin{figure}[ht]
%    \centering
%    \includegraphics[width=1.\linewidth]{latex/image/we_vs_others.pdf}
%    \caption{Comparison of our models with the baselines.}
%    \label{fig:data}
%\end{figure}

%\begin{figure}[ht]
%    \centering
%    \includegraphics[width=1.\linewidth]{latex/image/EPE_length_seq.pdf}
%    \caption{End-point-error depending on the length of the input sequence for AMDNet-RNN.}
%    \label{fig:tr}
%\end{figure}

\subsection{Simulator Dataset Results}

We evaluate our method and FlowNet3D in the simulated environment. Results in Table \ref{tab:simulator} demonstrate that our method can generalize well to simulated data. Moreover, it has a very accurate static/dynamic segmentation. It allows us to achieve extremely high quality on the static objects.

\begin{table}[]
\centering
\begin{tabular}{|c|c|c|}
\hline
Method     & 3D EPE & 3D EPE-dynamic \\ \hline
AMDNet-RNN & 0.0174 & 0.1611         \\ \hline
FlowNet3D  & 0.1533 & 0.1671         \\ \hline
\end{tabular}
\caption{Comparison results on the simulator dataset without road surface.}
\label{tab:simulator}
\end{table}

\begin{figure}[ht]
    \centering
    \begin{subfigure}[b]{0.425\textwidth}
        \includegraphics[width=\textwidth]{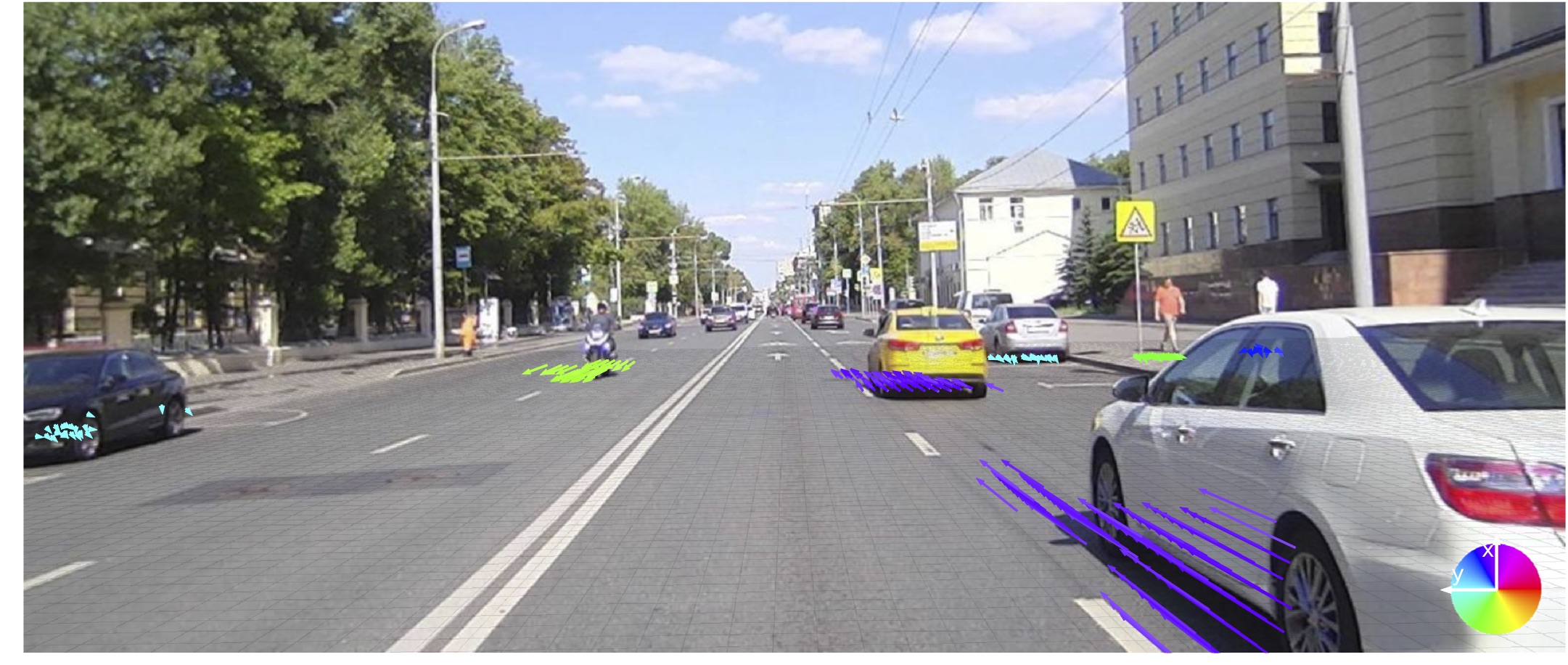}
    \end{subfigure}
    \begin{subfigure}[b]{0.425\textwidth}
        \includegraphics[width=\textwidth]{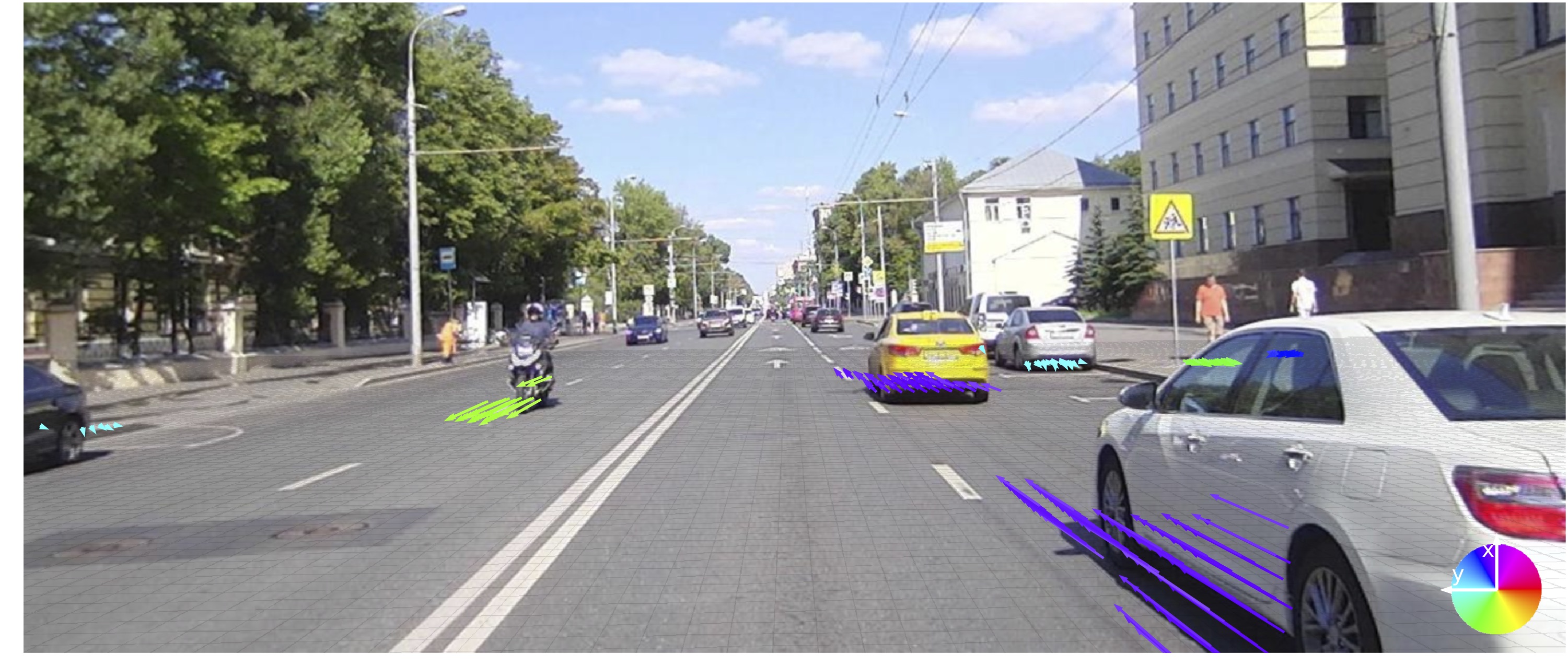}
    \end{subfigure}
    \begin{subfigure}[b]{0.425\textwidth}
        \includegraphics[width=\textwidth]{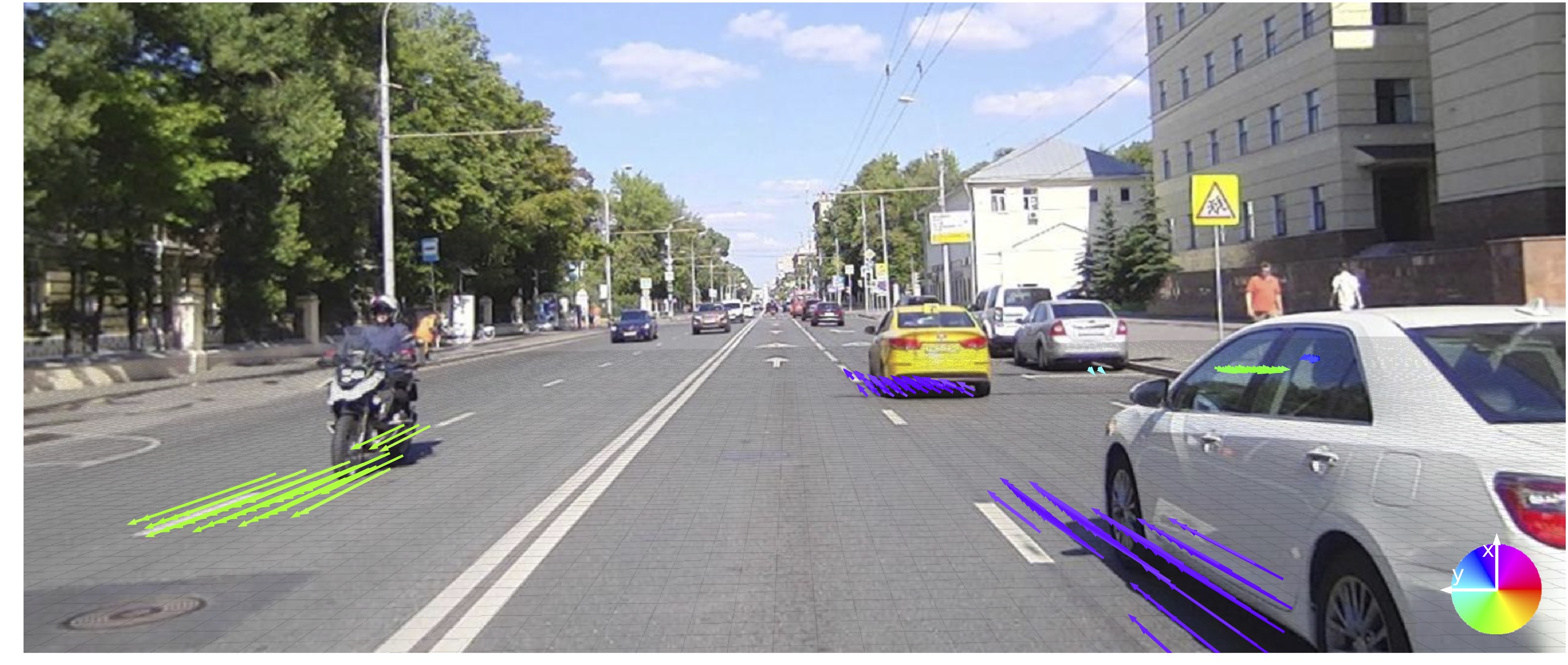}
    \end{subfigure}
    \caption{Motion estimation of common road participants. Though biker is not presented in the training data, their speed is also estimated precisely. To visualize the output, we project the bird's-eye view grid into the image and illustrate the speed of moving cells with arrows. The lengths of the arrows are proportional to the speed; the direction of the speed is  shown with the arrows' and color-coded in HSV color space.}
    \label{fig:moscow}
\end{figure}

\subsection{Post-processing with DBSCAN}

\begin{figure}[ht]
    \centering
    \includegraphics[width=0.9\linewidth]{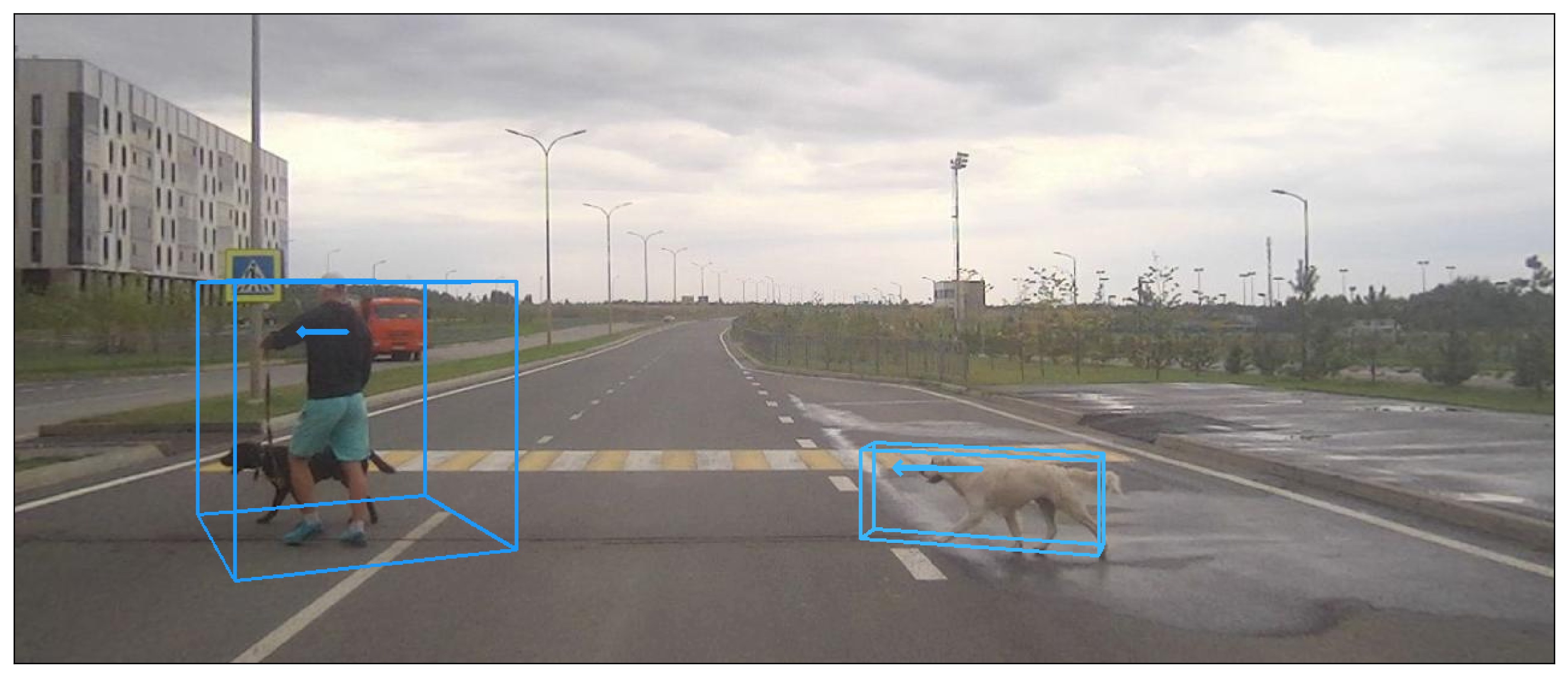}
    \caption{Example of output post-processing with DBSCAN. We place bounding boxes over clusters and estimate velocity as the mean velocity inside the cluster.}
    \label{fig:dog_bbox}
\end{figure}

In applications, it can be more convenient to work with bounding boxes than with grids. It can simplify tracking and prediction.  We propose a simple technique to obtain bounding boxes for dynamic objects. We binarize the segmentation grid by some threshold and then convert non-zero cells into vectors of the following form $(x, y, v_x, v_y)$. Where $x$ and $y$ are coordinates of the cell in the local coordinate system and  $v_x, v_y$ are velocities of the cell. Then we run the DBSCAN \cite{dbscan} algorithm to cluster points and finally, we find minimal bounding box for each cluster in the bird-eye-view. 
We estimate the height of the box as the coordinate of the highest LiDAR point inside the box. Figure \ref{fig:dog_bbox} demonstrates output bounding boxes. DBSCAN can set boxes on the stand-alone objects very accurately, whereas large groups of dynamic objects with similar velocity vectors can be grouped in the same cluster.

\section{Conclusion}

In this work we propose a novel architecture for 2D velocity estimation and dynamic/static segmentation. While previous works are not real-time and rely on the point clouds preprocessing, we achieve high-performance and real-time inference with the end-to-end approach. We propose a novel ego-motion compensation layer which allows taking ego-motion into account without point cloud transforms. Our approach demonstrates state-of-the-art results on the KITTI  dataset while being trained on another dataset. We also qualitatively demonstrate the generalization ability of our method with examples of unusual dynamic objects.
Future work may include the addition of new data sources to improve speed estimation e.g. radar data or RGB optical flow.

{\small
\bibliographystyle{ieee}
\bibliography{ms}
}

\end{document}